\documentclass[letterpaper, 10 pt, conference]{ieeeconf}  

\IEEEoverridecommandlockouts                              

\usepackage{cite}
\usepackage{amsmath,amssymb,amsfonts,empheq}
\usepackage{mathtools}
\usepackage{textcomp}
\usepackage{url}
\usepackage{caption}
\usepackage{floatrow}   
\usepackage{subcaption}
\usepackage{graphicx}
\usepackage[framemethod=tikz]{mdframed}
\usepackage{tablefootnote} 
\usepackage{algorithm}
\usepackage{algpseudocode}
\usepackage{multirow}
\usepackage{titlecaps}

\makeatletter
\let\NAT@parse\undefined
\makeatother
\usepackage{hyperref}

\newcommand\NetworkName{StereoVoxelNet}
\newcommand\CostVolumeName{voxel cost volume}

\title{\LARGE \bf
\NetworkName{}: Real-Time Obstacle Detection Based on Occupancy Voxels from a Stereo Camera Using Deep Neural Networks
}

\author{Hongyu Li$^{1,2\dagger}$, Zhengang Li$^{3*}$, Ne\c{s}et \"{U}nver Akmandor$^{1,3*}$, Huaizu Jiang$^{2}$, Yanzhi Wang$^{3}$, and Ta\c{s}k{\i}n Pad{\i}r$^{1,3}$
\thanks{This research is supported by the National Science Foundation under Award Number 1928654.}
\thanks{$^{1}$Institute for Experiential Robotics, Northeastern University}
\thanks{$^{2}$Khoury College of Computer Sciences, Northeastern University}
\thanks{$^{3}$Department of Electrical and Computer Engineering, Northeastern University, Boston, MA, 02115, USA}%
\thanks{$^{*}$Equally contributed authors}%
\thanks{$^{\dagger}${\tt\small li.hongyu1@northeastern.edu}}%
}
\begin{document}

\maketitle
\thispagestyle{empty}
\pagestyle{empty}

\newcommand{\hj}[1]{\textcolor{magenta}{HJ: #1}}
\newcommand{\nua}[1]{\textcolor{cyan}{NUA: #1}}
\newcommand{\lhy}[1]{\textcolor{red}{LHY: #1}}

\begin{abstract}

Obstacle detection is a safety-critical problem in robot navigation, where stereo matching is a popular vision-based approach. While deep neural networks have shown impressive results in computer vision, most of the previous obstacle detection works only leverage traditional stereo matching techniques to meet the computational constraints for real-time feedback. This paper proposes a computationally efficient method that employs a deep neural network to detect occupancy from stereo images directly. Instead of learning the point cloud correspondence from the stereo data, our approach extracts the compact obstacle distribution based on volumetric representations. In addition, we prune the computation of safety irrelevant spaces in a coarse-to-fine manner based on octrees generated by the decoder. As a result, we achieve real-time performance on the onboard computer (NVIDIA Jetson TX2). Our approach detects obstacles accurately in the range of 32 meters and achieves better IoU (Intersection over Union) and CD (Chamfer Distance) scores
with only 2\% of the computation cost of the state-of-the-art stereo model. Furthermore, we validate our method's robustness and real-world feasibility through autonomous navigation experiments with a real robot. Hence, our work contributes toward closing the gap between the stereo-based system in robot perception and state-of-the-art stereo models in computer vision. To counter the scarcity of high-quality real-world indoor stereo datasets, we collect a 1.36 hours stereo dataset with a mobile robot which is used to fine-tune our model. The dataset, the code, and further details including additional visualizations are available at \href{https://lhy.xyz/stereovoxelnet/}{https://lhy.xyz/stereovoxelnet/}.

\end{abstract}

\section{INTRODUCTION}

\label{sec:intro}
	
    To safely navigate in dynamic and unknown environments, robots need to perceive and avoid obstacles in real-time, usually with a limited financial or computation budget.
    Among 3D perception sensors, stereo camera systems have been widely used due to their lightweight design and low cost. However, performing depth estimation with the stereo camera is nontrivial; we need to balance efficacy and efficiency. Therefore, most stereo-based navigation approaches sacrifice the estimation accuracy by using the traditional methods or reducing image resolution to provide real-time feedback \cite{7152384, 6696922, 7354095, 9340699}.

    \begin{figure}[ht!]
    \centering
    \begin{subfigure}{.49\linewidth}
        \centering
        \includegraphics[width=\linewidth]{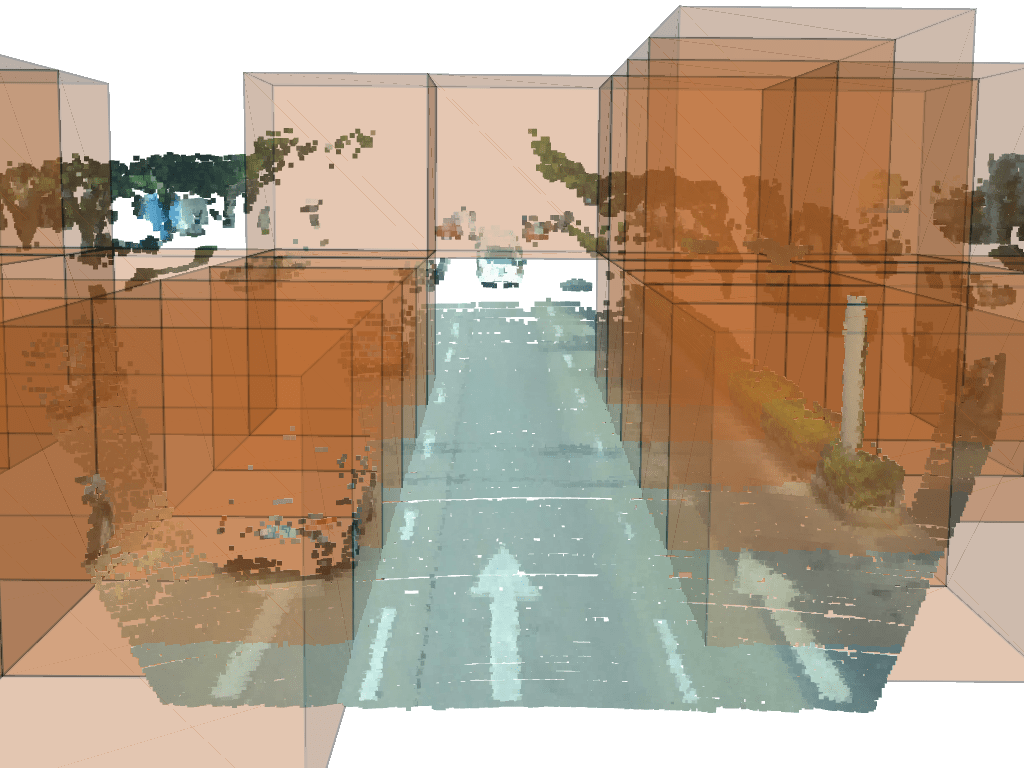}
        \caption{Level 1}
    \end{subfigure}
    \begin{subfigure}{.49\linewidth}
        \centering
        \includegraphics[width=\linewidth]{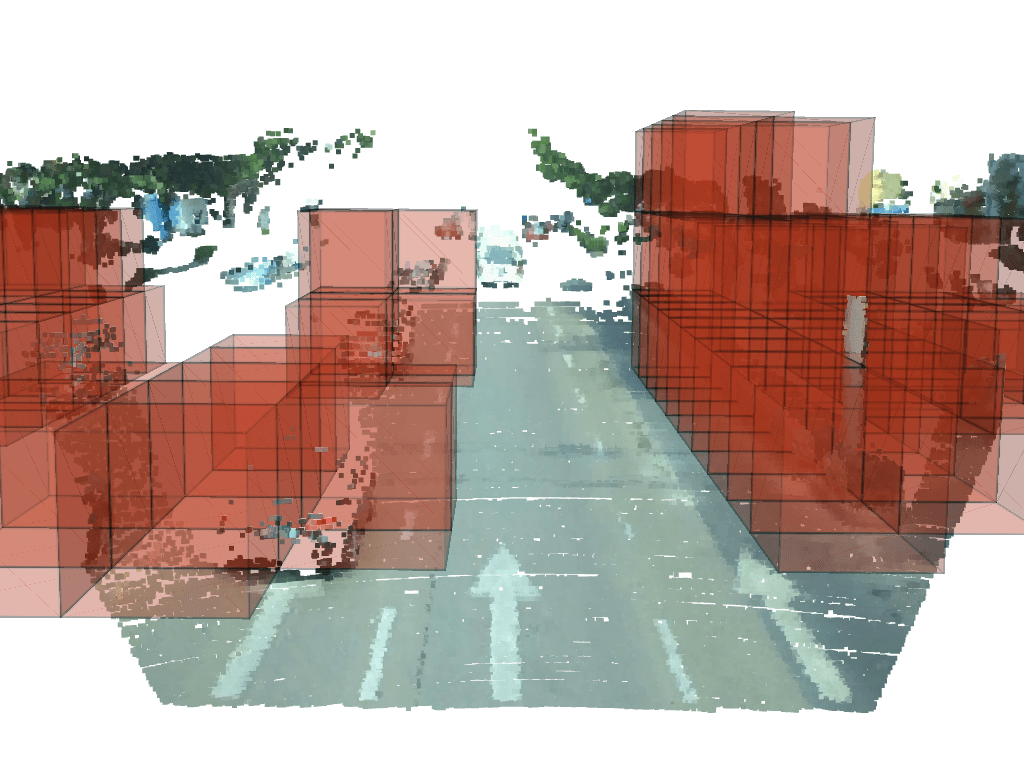}
        \caption{Level 2}
    \end{subfigure}
    \begin{subfigure}{.49\linewidth}
        \centering
        \includegraphics[width=\linewidth]{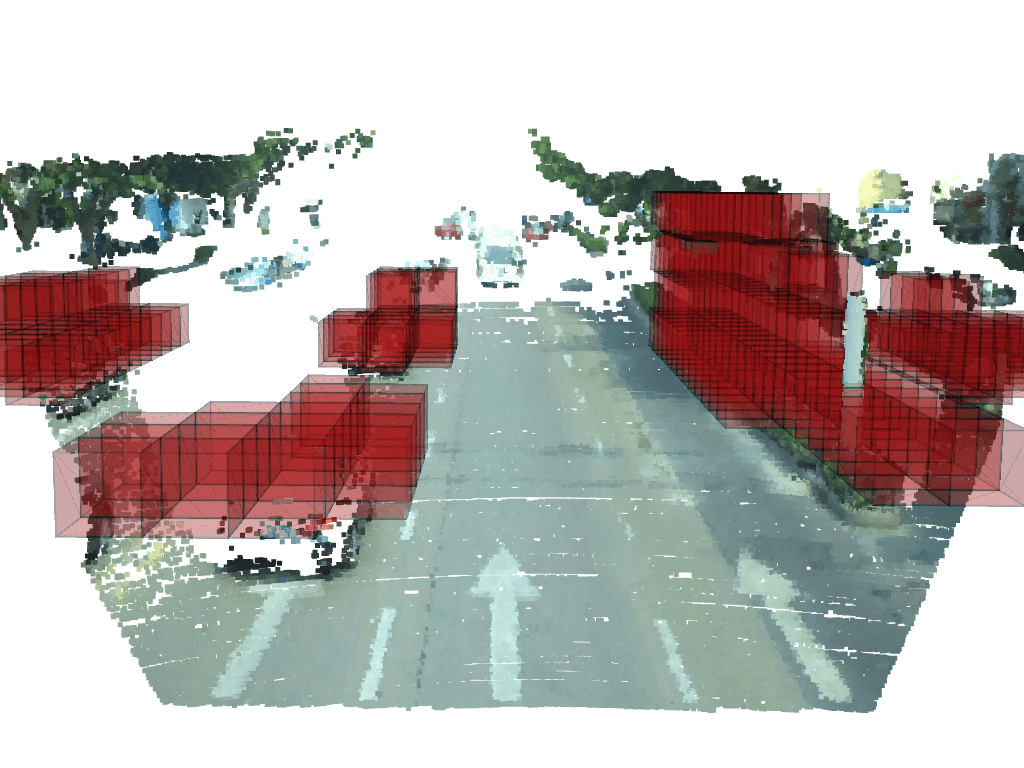}
        \caption{Level 3}
    \end{subfigure}
    \begin{subfigure}{.49\linewidth}
        \centering
        \includegraphics[width=\linewidth]{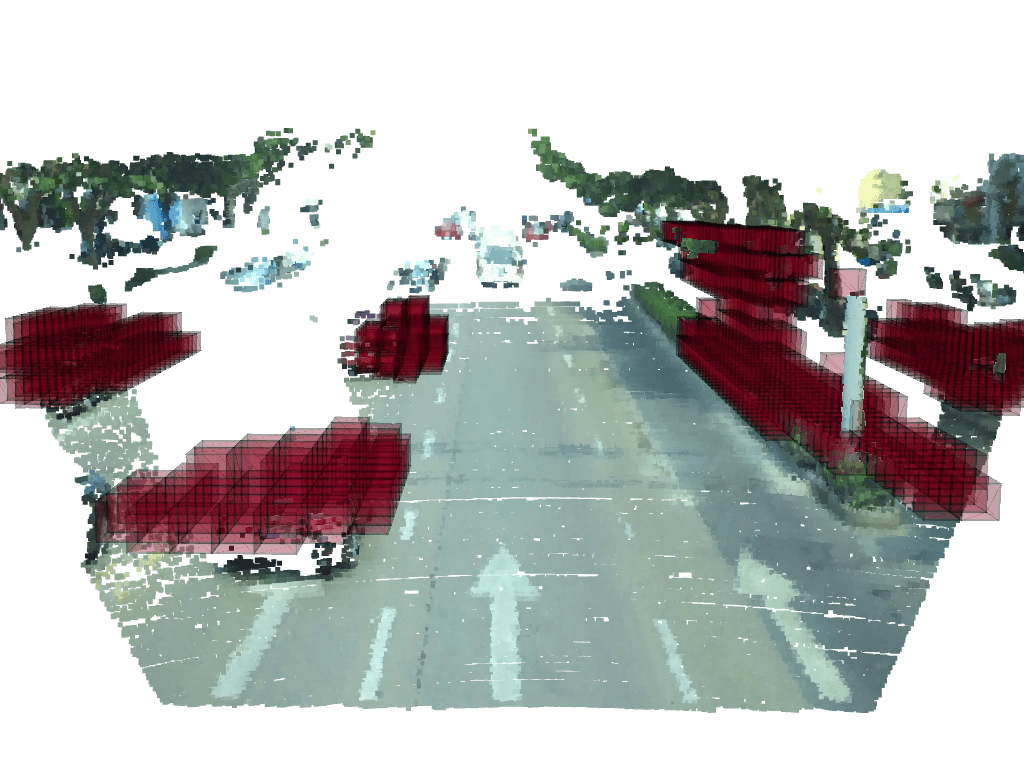}
        \caption{Level 4}
    \end{subfigure}

    \hfill
    
    \caption{Hierarchical output.~\NetworkName{} generates voxels from a stereo pair\cite{yang2019drivingstereo} to represent the detected location of the obstacles in the range of 32 meters in a coarse-to-fine grained manner (from top left to bottom right). The resolutions of the voxel grids are $8^3, 16^3, 32^3,$ and $64^3$, respectively.}
    \label{fig:voxel-sample}
    \end{figure}
    
  
    
    Deep learning stereo models have shown promising results in depth estimation\cite{MIFDB16}, however, there are mainly two limitations to their usability in practical and agile robot navigation. Firstly, even the lightweight deep learning stereo model \cite{shamsafar2022mobilestereonet} produces inadmissible latency. This is because the dense depth information 
    produced is redundant for obstacle avoidance tasks during motion planning.     
    In most stereo matching procedures, the depth of each pixel is estimated,
    and most of the individual points are superfluous in close proximity, thus discretized voxels are used to represent the occupancy information in typical robot applications \cite{7152384, 6696922, 7354095, 9340699, akmandor_reactive_2021, akmandor_deep_2022} that require fast computation.
    
    
    
    Secondly, the current state-of-the-art stereo models lack robustness. Even though they can achieve high accuracy in disparity estimation (2D), they are prone to generate more artifacts being transformed in 3D. The artifact is typically around the object boundaries and at a far distance \cite{9562056, Weng_2019_ICCV_Workshops}. After the transformation, the artifacts are embodied in the long-tail form \cite{Weng_2019_ICCV_Workshops}, due to their pipeline's objective bias. Consequently, an extra outlier filtering is needed to recast the adequate point cloud for robot navigation \cite{9340699}. Such a procedure often requires careful tuning of parameterization and lacks robustness.
    
    To address the aforementioned shortcomings, we propose an approach that detects the obstacles efficiently and accurately, leveraging the deep learning stereo model. Unlike existing models that estimate stereo disparity (and then point cloud) from stereo images, our novel end-to-end network directly outputs the occupancy by resorting to a voxel grid, reducing the redundancy of the point cloud. We further improve both the accuracy and the computation efficiency of the interlacing cost volume \cite{shamsafar2022mobilestereonet} by ignoring the obstacles that are far away and avoiding unnecessarily high granularity. Such a design also reduces the noises introduced by the coordinate transformation from the stereo disparity to the 3D point cloud. Our solution can detect obstacles accurately in the range of 32 meters (See Section \ref{sec:experiment}) which clearly outperforms the off-the-shelf sensors like Intel RealSense and Microsoft Kinect.

    
    Furthermore, taking advantage of the octree structure and sparse convolution operations, we predict the occupancy of obstacles in a coarse-to-fine grained manner (Fig. \ref{fig:voxel-sample}). This also saves computational resources by eliminating the need for further processing the confident unoccupied spaces which are irrelevant to the navigation task at early stages.
    Hence, we achieve real-time inference for obstacle detection using an onboard computer. Our approach contributes to closing the gap between computer vision and robot perception.

    Over the past decade, numerous autonomous driving stereo datasets have been constructed, such as KITTI \cite{KITTI2012}, DrivingStereo \cite{yang2019drivingstereo}, nuScenes \cite{caesar_nuscenes_2020}, and Waymo \cite{Sun_2020_CVPR}, however, stereo models trained on these datasets typically perform inadequately in indoor environments. To address the diversity in depth, prior works \cite{shamsafar2022mobilestereonet, liu_local_2022} utilize the synthetic SceneFlow \cite{MIFDB16} stereo dataset, but they encounter performance degradation in the real-world applications. Consequently, we construct a stereo dataset containing both indoor and outdoor environments via robot teleoperation and incorporate data from onboard sensors including IMU, stereo camera, and a 32-beam LiDAR. The details of the dataset are presented in Section \ref{sec:dataset}.
    
    The contributions of this paper can be summarized as follows:
    \begin{itemize}
        \item We design and implement a novel deep neural network that takes stereo image pairs as input and produces occupancy voxels through an efficient cost volume formulation strategy in real-time using an onboard computer. 
        \item We exploit the sparsity by performing optimization through octree structure and sparse convolution and saving computation resources on confident unoccupied space. We further integrate our perception model with the motion planner module, thus achieving adaptive perception on demand \cite{dean1988analysis, liu2021anytime, Chitta_2020_WACV} during navigation.
        \item We construct a real-world stereo dataset that contains both indoor and outdoor scenes, collected by robot teleoperation. Our contributed dataset complements the existing datasets like JRDB \cite{martin-martin_jrdb_2021} to enhance the dataset scale and availability for mobile robot navigation.
    \end{itemize}

\section{Related Work}
\label{sec:related}

\textbf{Obstacle detection} using the stereo camera is typically approached by depth estimation, utilizing the classical stereo matching algorithms, such as Semi-global Matching (SGM) \cite{sgm}. The majority of works construct a pipeline that first estimates the depth map and then extracts the obstacles \cite{6696922, 7152384, 7354095, 9340699, doi:10.1126/scirobotics.abg5810}. However, such a pipeline lacks efficiency since the depth of every pixel is estimated by the stereo matching module. On the contrary, obstacles typically account for only a small portion of pixels.
Pushbroom Stereo \cite{barry2014pushbroom}  tackled the inefficiency by performing block matching sparsely, which is capable of running at 120Hz onboard, but it suffers from high false negatives due to the lack of global attention. 

Deep learning end-to-end models are also attempted by several works. \cite{7358076} developed an end-to-end classification model to perform trail perception and navigation. Similarly \cite{bojarski2016end} and DroNet\cite{dronet} proposed to use an end-to-end CNN model for the output steering angle. Since this kind of network model learns obstacle avoidance intrinsically, their performance is sensitive to the training scene and they tend to have poor domain adaptation.


Our work in this paper focuses on the sub-modules of visual navigation as in \cite{9340699, doi:10.1126/scirobotics.abg5810} and could improve the overall system by replacing their depth estimation-related modules with our proposed approach. \\

\textbf{Stereo matching} is used to perform depth estimation by measuring the disparity between stereo images. The classical method like SGM has been widely used in robot perception\cite{6696922, 7152384, 9340699, doi:10.1126/scirobotics.abg5810}. As in various computer vision tasks, the performance of the traditional approach has been surpassed by the recent deep learning models and often by a large margin, which also applies to stereo matching tasks \cite{10.5555/2946645.2946710, 8237440}.
Recent stereo matching work has been focused on developing an end-to-end network to predict the disparity map from a stereo pair \cite{MIFDB16, shamsafar2022mobilestereonet, xu2022ACVNet}. In such an end-to-end network, the matching procedure is often performed by cost volume construction and cost volume aggregation. A cost volume $C$ is used to correlate two unary feature maps generated by the input stereo pair passing through the feature extractor. \cite{7780983, shamsafar2022mobilestereonet} utilize 3D cost volume
formulated as
\begin{align}\label{eqn:3d_cost_vol_baseline}
C(d,x,y) = F(f_L(x,y),f_R(x-d,y)),
\end{align}
where $F$ is a function that measures the correlation between the left and right feature maps. $d$ is within the range of $(0,d_{max})$, where $d_{max}$ is a designated maximum disparity level and typically set to 192 in recent works. 

DispNet \cite{MIFDB16} has reached a significantly higher performance compared with SGM on multiple datasets since 2015. Nevertheless, the deployment of recent deep learning stereo models in mobile robotics is still unpopular. In fact, it is a huge challenge to run deep learning stereo models in real-time using an onboard computer. \\


\textbf{Octree representation} has been widely used in the field of robotics \cite{hornung13auro, labbe_rtab-map_2019}. Our octree decoder design is related to two seminal works: Octree Generating Network \cite{Tatarchenko_2017_ICCV} and Quadtree Generating Network \cite{Chitta_2020_WACV}. While the first one performs 3D reconstruction, the latter work is on 2D image segmentation. Different from these works, we contribute to the adaptation to stereo upstream and its integration with the robot navigation module to achieve adaptive perception on demand, where the perception granularity can automatically be determined by the robot's demand for obstacle avoidance.

\section{\NetworkName{}}
\label{sec:method}

\begin{figure}[!ht]
\centering
\includegraphics[width=\linewidth]{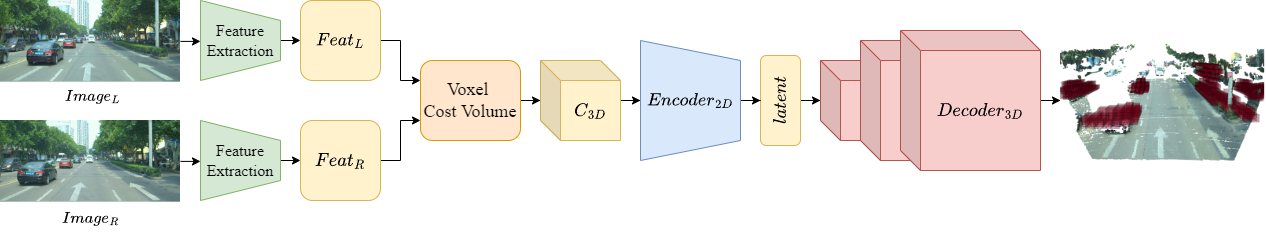}
\caption{Our proposed network, \NetworkName{}, takes a stereo pair as input and produces a voxel occupancy grid.}
\label{fig:voxelnet}
\end{figure}
	
\begin{figure*}[!ht]
\centering

\vspace*{0.15cm}

\includegraphics[width=0.85\linewidth]{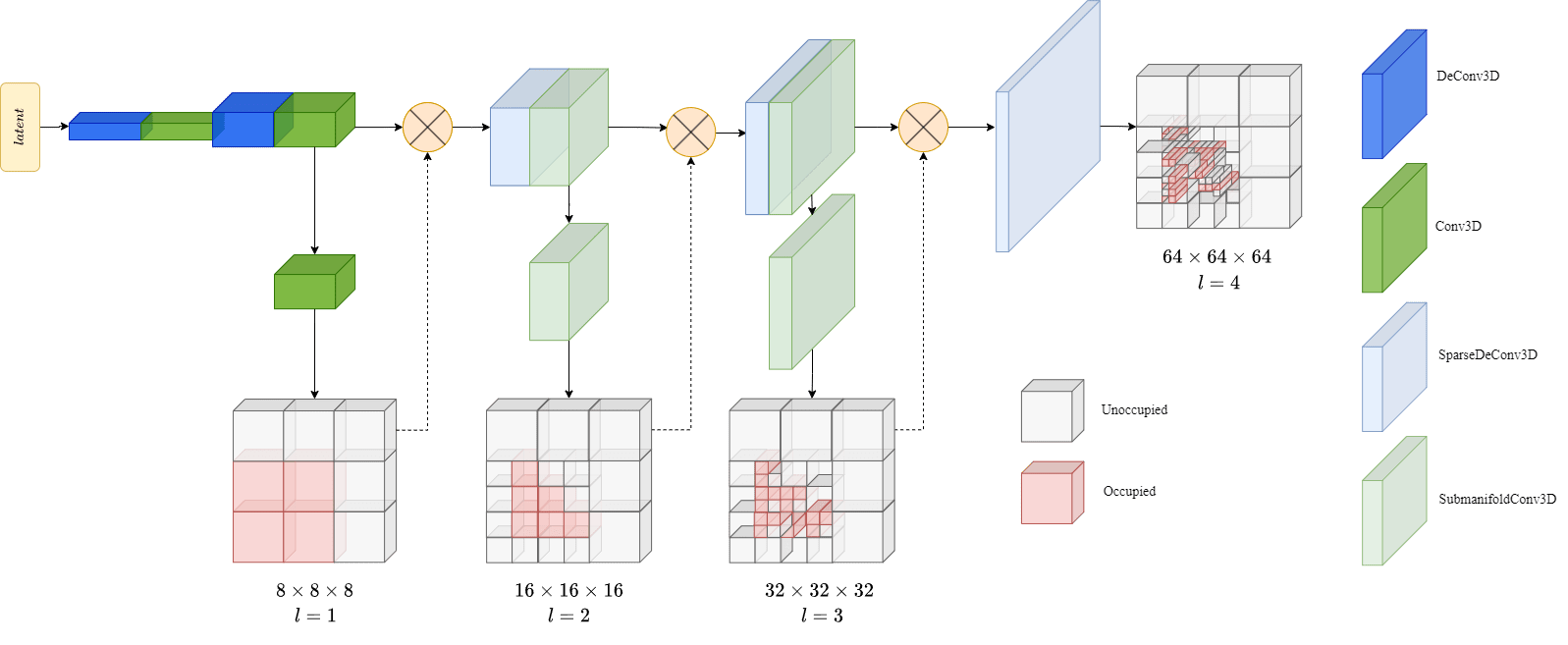}
\caption{Our octree decoder takes the latent vector as input and produces coarse-to-fine grained output. The output of each level is used as a mask to prune the feature matrix for the next level. We set the threshold as 0.5 to binarize the mask. Thus, the confident unoccupied spaces become zeros and are pruned using sparse convolutions.}
\label{fig:decoder}
\end{figure*}

Inspired by stereo matching models \cite{shamsafar2022mobilestereonet, xu2022ACVNet, cheng2020hierarchical, 10.5555/2946645.2946710}, we design our end-to-end network, \NetworkName{}, by integrating three main components as depicted in Fig. \ref{fig:voxelnet}: \emph{i}) feature extraction module for unary feature extraction, \emph{ii}) \CostVolumeName{}: a lightweight cost volume to combine unary features, and \emph{iii}) an encoder-decoder structure for voxels generation. The first two parts extract and aggregate features, while the last part predicts voxels from the low-level feature volume base on an octree structure. 

Similar to the navigation framework \cite{akmandor_reactive_2021}, we define a 3D grid composed of $N$ voxels with respect to the robot's coordinate frame. We denote each voxel as a cubic volume with a side length $l_{v}$. The 3D grid $N=n_x \times n_y \times n_z$, where $n_x$, $n_y$, and $n_z$ represent the number of voxels on $x$-axis, $y$-axis, and $z$-axis, respectively. Therefore, the physical distance $s_{\{x,y,z\}}$ in each dimension can be calculated as $n_{\{x,y,z\}} \times l_v$. We refer to this volumetric voxel grid structure as our region of interest (ROI).

\textbf{Feature Extraction: }
We utilize a simplified feature extraction module by reducing the number of channels in MobileStereoNet\cite{shamsafar2022mobilestereonet}. A feature map of size $C \times \frac{H}{4} \times \frac{W}{4}$ is produced from the feature extraction module. $C$ is the number of feature channels, and $H, W$ is the image's height and width, respectively. 

\textbf{\titlecap{\CostVolumeName{}}: }
We leverage \CostVolumeName{} to aggregate unary features efficiently. A detailed explanation of \CostVolumeName{} is given in Section \ref{sec: VCV}.

\textbf{Encoder-Decoder: }
We use an encoder-decoder component to produce voxels\cite{10.1007/978-3-030-11009-3_37, choy20163d, 10.5555/3157096.3157287, 10.5555/3294771.3294823}, whereas the encoder is a collection of 2D convolutional layers that extracts a latent vector from the feature volume, and the decoder is composed of a set of 3D deconvolutional layers with an octree generating structure. To predict obstacle occupancy with higher efficiency, we optimize the decoder component by exploiting the sparsity of the occupancy grid and applying sparse convolution. A detailed explanation of our octree optimization is in Section \ref{sec: octree_decoder}.

\subsection{\titlecap{\CostVolumeName{}}}
\label{sec: VCV}


In most recent stereo models, the cost volume is one of the most computationally demanding modules. Furthermore, its output dimension linearly affects the computation cost of the consecutive modules. To improve the efficiency of the entire network, we propose a voxel cost based on the interlacing cost volume\cite{shamsafar2022mobilestereonet} with a consideration of the physical structure of the final output - voxel occupancy grid. We eliminate redundant and irrelevant disparity information. Instead of continuously traversing $\{d_i \mid i=1,2,\dots,d_{max}\}$ in equation (\ref{eqn:3d_cost_vol_baseline}) \cite{shamsafar2022mobilestereonet}, we replace it with $D_{vox}$, a subset of disparity levels.
\begin{align}\label{eqn:3d_cost_vol}
C(d,x,y) = Interlace\{f_L(x,y),f_R(x-d_{vox},y)\},
\end{align}
where $d_{vox} \in D_{vox}$. $Interlace$ is an operator proposed by \cite{shamsafar2022mobilestereonet}, where the left feature map and right feature map of size $C \times H \times W$ are aggregated one-by-one into $2C \times H \times W$. Since $d$ and $z$ have an inverse relationship, if we traverse continuously on the disparity dimension with a constant gap, we get a skewed distribution of the respective depth $z$. Particularly, we retrieve sparser features at a farther distance and denser features as getting closer. We use the voxel size $l_{v}$ as a gap, traversing $z$ in the range of $(0, z_{max})$ where $z_{max} = n_z \times v_{size}$. Consequently, we get $|D_{vox}| = n_z$. 

Our algorithm is presented in Algorithm \ref{alg:vcv}. The \CostVolumeName{} comes with two functionalities: \emph{i}) It prunes the information which is not in our ROI. \emph{ii}) It enables a sparser sampling of disparity without harming the overall accuracy.

\renewcommand{\algorithmicrequire}{\textbf{Input:}}
\renewcommand{\algorithmicensure}{\textbf{Output:}}
\begin{algorithm}
\caption{\titlecap{\CostVolumeName{}}}\label{alg:vcv}
\begin{algorithmic}
\Require $f_u$, $b$, $z_{max}$, $\alpha$, $l_{v}$
\Ensure $C$
\State $c \gets f_u \times b$   \Comment{constant for each stereo camera}
\State $D_{vox}=\emptyset$
\For{$z=0$ to $z_{max}$ with step size of $\alpha \times l_v$ }
    \State $D_{vox} = D_{vox} \cup \{\frac{c}{z}\}$
\EndFor
\For{$d_{vox}$ in $D_{vox}$}
    \State $C(d,x,y) \gets Interlace\{f_L(x,y),f_R(x-d_{vox},y)\}$
\EndFor
\end{algorithmic}
\end{algorithm}

In the algorithm, $f_u$ represents the focal length, $b$ represents the baseline of the stereo camera. $\alpha$ is a constant controlling the traversing step size, and $l_v$ is the size of a voxel as mentioned.

\subsection{Octree Decoder}
\label{sec: octree_decoder}
An octree is a tree-like data structure where each node has eight children \cite{meagher1982geometric}. Like binary tree (1D) and quadtree (2D), octree (3D) is a compact representation of the space that advances in both computation efficiency and storage.
Benefiting from its tree-like data structure, an octree is naturally hierarchical. We can represent each level of the octree as $l=\{ 1,2,\dots,L_{level}\}$. For the level $l$, we could obtain the dimension of output $N_l =\delta \times (2^l \times 2^l \times 2^l)$, where $\delta$ is the initial resolution. Having a fixed physical size $s^3$ of the 3D grid, the output of our model follows a coarse-to-fine grained pattern.

Unlike \cite{Tatarchenko_2017_ICCV} where they define each voxel of output voxel grid $y$ at the location $p \in \mathbb{R}^3$ and octree level $l$ as being one of three states: \emph{empty}, \emph{filled}, or \emph{mixed}, we simplify it into two states \emph{occupied} (1) and \emph{unoccupied} (0) (Fig. \ref{fig:decoder}).

\begin{equation}\label{eqn:octree_state}
y_{l,p}=
\begin{cases}
&0 \text{ if all the children are unoccupied},\\
&1 \text{ if any of the children is occupied}.
\end{cases}
\nonumber
\end{equation}

To use a network to predict the occupancy label for each voxel, we feed the latent vector through a set of deconvolution and convolution layers to obtain the first coarse output $y_{1,p}$. We employ occupancy label $y_{1,p}$ as a mask to sparsify the reconstructed feature using element-wise multiplication. The process is repeated until the desired resolution is reached.

To calculate the loss of the octree, we calculate the weighted sum of the negative Intersection of Union (IoU) loss of each tree level. 
\begin{equation}\label{eqn:level_loss}
\mathcal{L}_l= \sum_{p} 1-IoU(y_{l,p}, \hat{y}_{l,p}),
\nonumber
\end{equation}
where $y_{l,p}$ and $\hat{y}_{l,p}$ correspond to the prediction and ground-truth label at level $l$. The overall loss is given by:
\begin{equation}\label{eqn:total_loss}
\mathcal{L} = \sum_l{w_l \mathcal{L}_l},
\nonumber
\end{equation}
where $w_l$ is a designated weight for each level. We set weights $w_1$ to $w_4$ as $[0.30, 0.27, 0.23, 0.20]$ empirically.

During inference, we can detect obstacles at the finest level $y_{L_{level}, p}$ in real-time utilizing the aforementioned methodologies. However, we further improve the efficiency using a small technique. Leveraging the hierarchical design of the octree decoder, we integrate our \NetworkName{} with the mobile robot navigation framework and achieve adaptive perception on demand. We use a simple heuristics function to model the octree level variation from the previous observation and permit early exit.

\begin{equation}
l^t=
\begin{cases}
&i-1~~~\text{if~} \sum_{p'} y_{i,p'}^{t-1}=0, \\
&i+1~~~\text{otherwise.}
\end{cases}
\end{equation}
where $l^0=L_{level}$, and we use the superscript $t$ to denote the time step. The heuristic function means if there is no obstacle detected in area $p'$ in front of the robot from the previous timestep $t-1$, the exit octree level $l$ of current timestep $t$ will decrease by one. In case of an obstacle is detected, the octree level will increase by one.


\section{Dataset Construction}
\label{sec:dataset}

\begin{figure}[ht!]
    \centering
    \includegraphics[height=4cm]{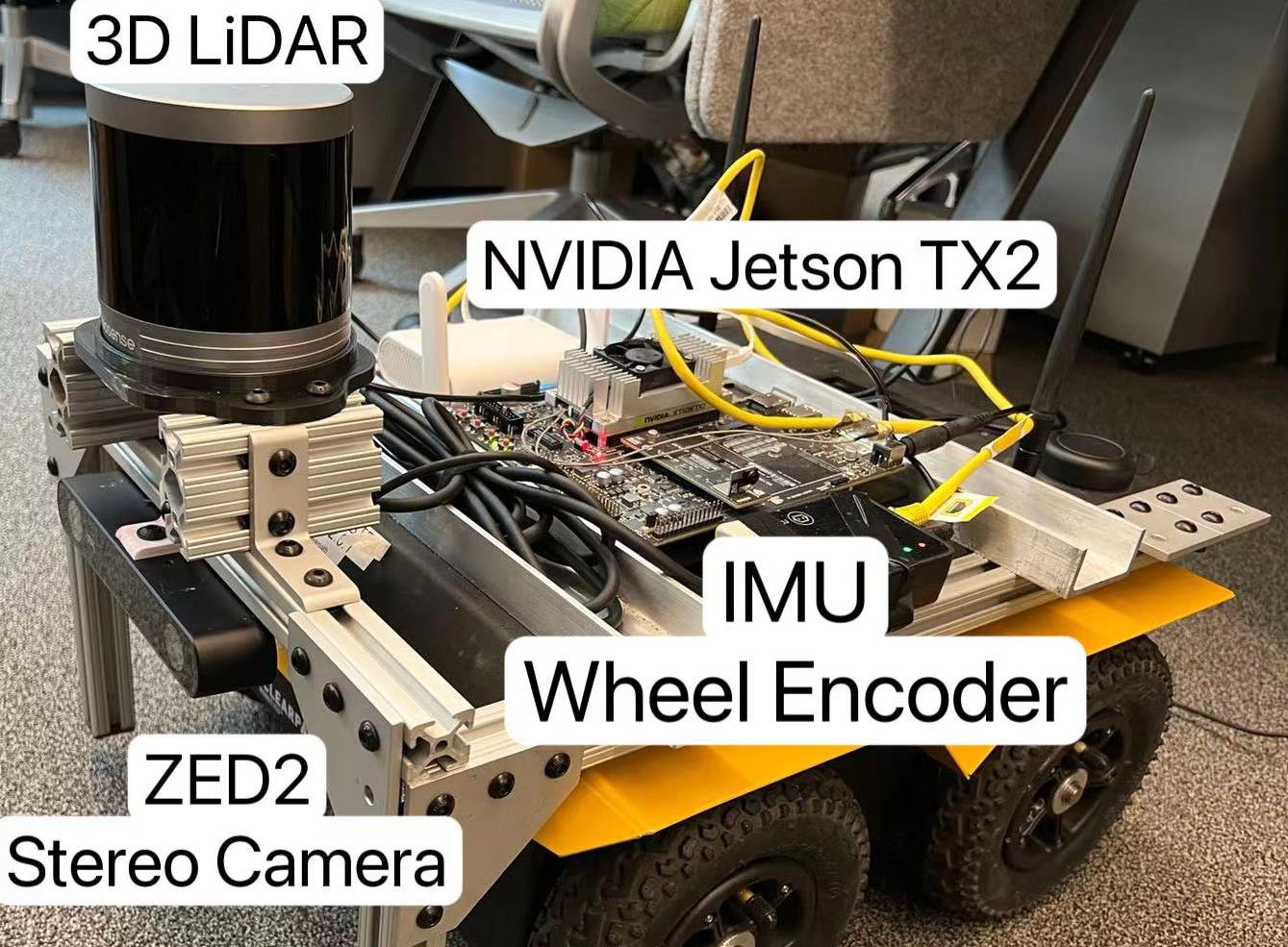}
    \caption{Our Jackal robot is teleoperated in indoor and outdoor environments carrying a suite of sensors to collect data.}
    \label{fig:jackal-robot}
\end{figure}

\begin{figure}[ht]
    \centering
    \begin{subfigure}{.5\linewidth}
        \centering
        \includegraphics[width=\linewidth, height=3cm]{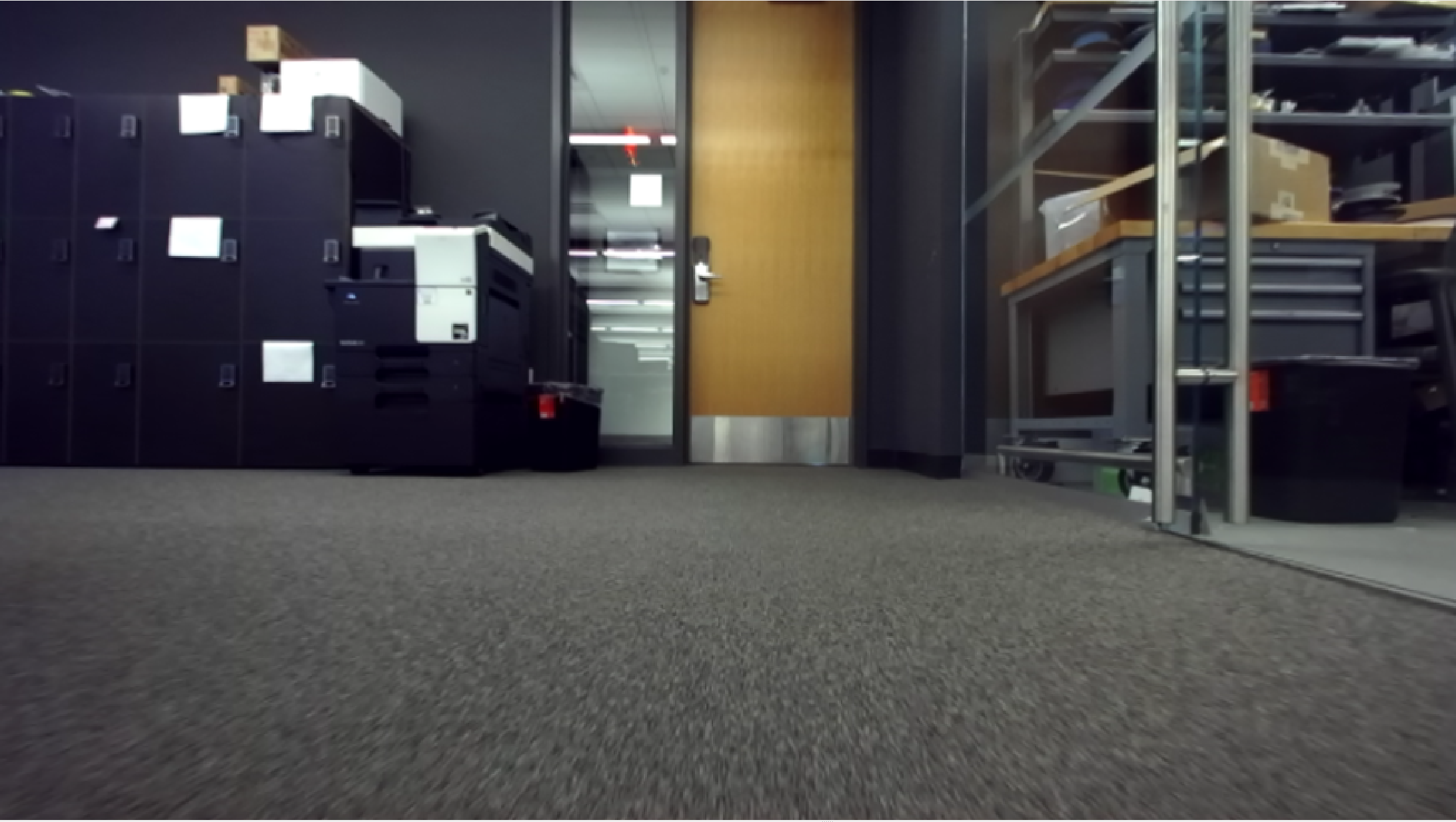}
    \end{subfigure}
    \hspace{-0.5cm}
    \begin{subfigure}{.5\linewidth}
        \centering
        \includegraphics[width=\linewidth, height=3cm]{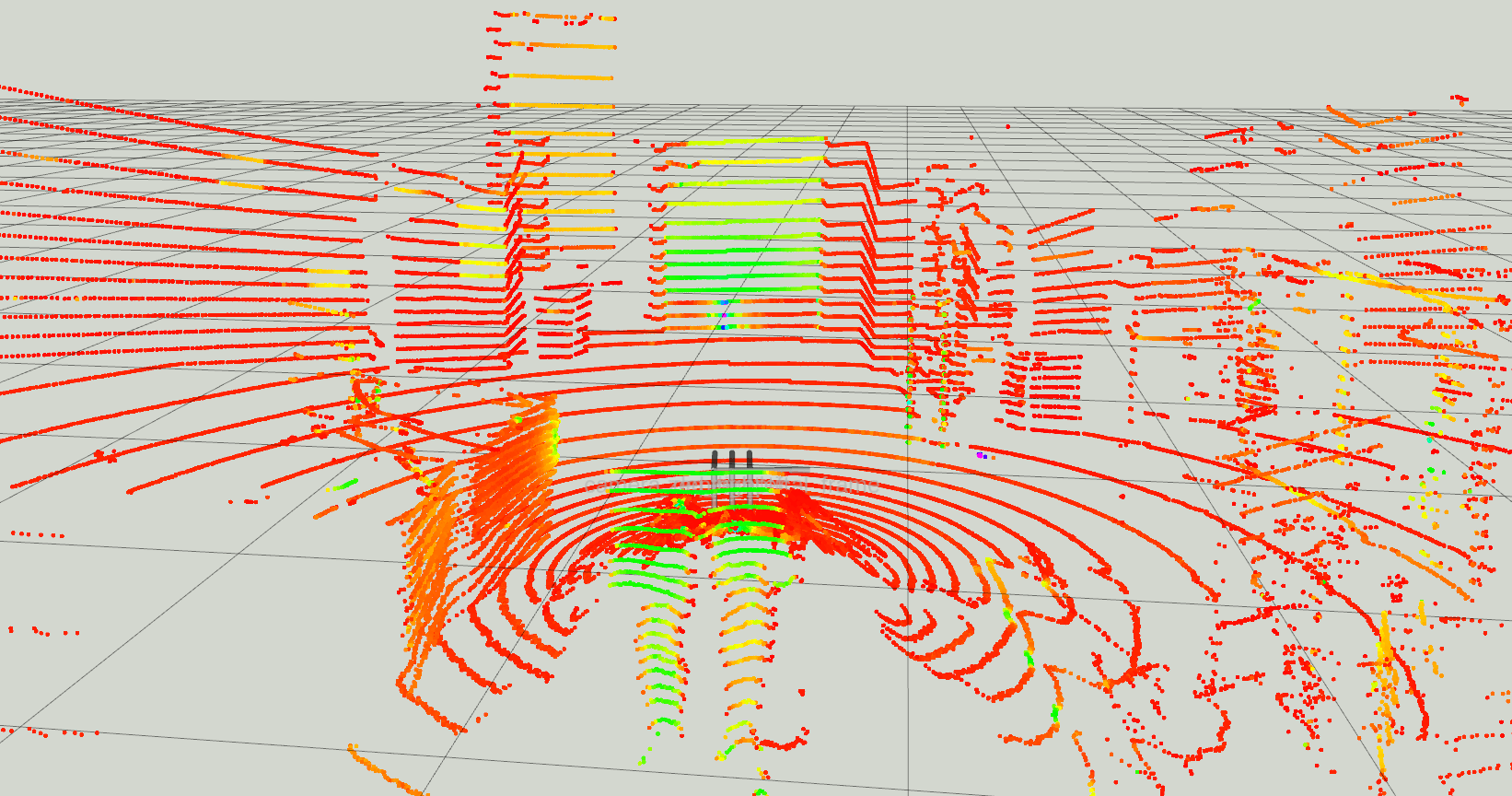}
    \end{subfigure}
    \caption{A sample from our dataset. The left image is rectified image captured by the ZED2 camera. The right image is 3D LiDAR point cloud visualized in RViz.}
    \label{fig:dataset}
\end{figure}

The dataset was collected using a Clearpath Robotics Jackal robot (Fig. \ref{fig:jackal-robot}) equipped with a front-facing ZED2 stereo camera and a RoboSense-RS-Helio-5515 32-Beam 3D LiDAR which operates at 10Hz.
All data is collected, synchronized, and stored in \emph{rosbag} format in ROS \cite{Quigley09}. 

The robot is teleoperated by a joystick controller to capture data from a variety of indoor and outdoor 
locations at the Northeastern University campus in Boston. These include different floors of the library and a research building, and the walking bridge. The dataset visualization is depicted in Fig. \ref{fig:dataset}. Different from social navigation datasets JRDB \cite{martin-martin_jrdb_2021} and SCAND \cite{karnan_socially_2022} where the robot navigates a crowded environment and imitates human behavior to avoid obstacles, our robot operates near static obstacles while navigating to include more samples of obstacles nearby.


\begin{figure*}[ht]
    \centering
    \begin{subfigure}{.19\linewidth}
        \centering
        \includegraphics[width=\linewidth, height=2.2cm]{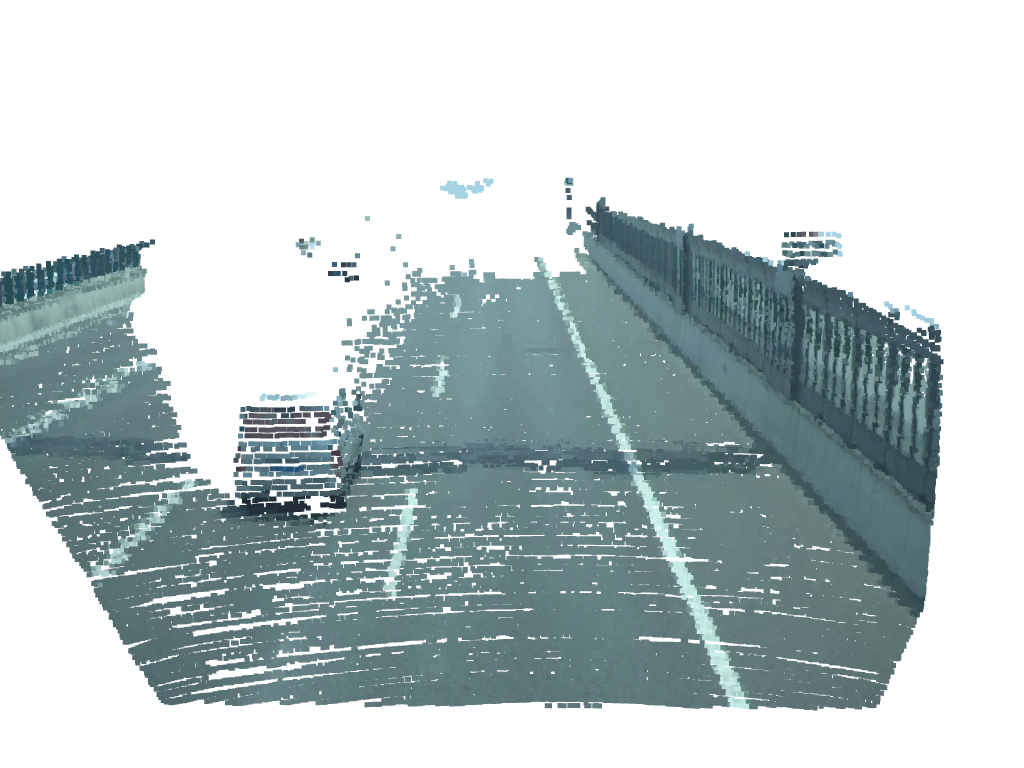}
    \end{subfigure}
    \begin{subfigure}{.19\linewidth}
        \centering
        \includegraphics[width=\linewidth, height=2.2cm]{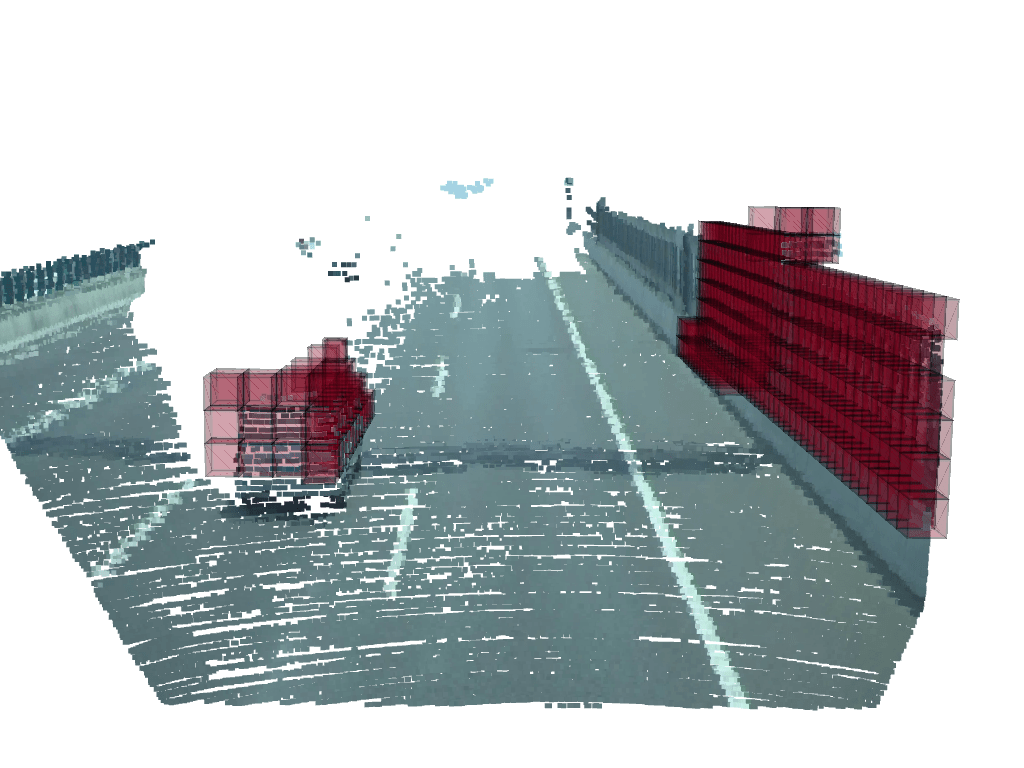}
    \end{subfigure}
    \begin{subfigure}{.19\linewidth}
        \centering
        \includegraphics[width=\linewidth, height=2.2cm]{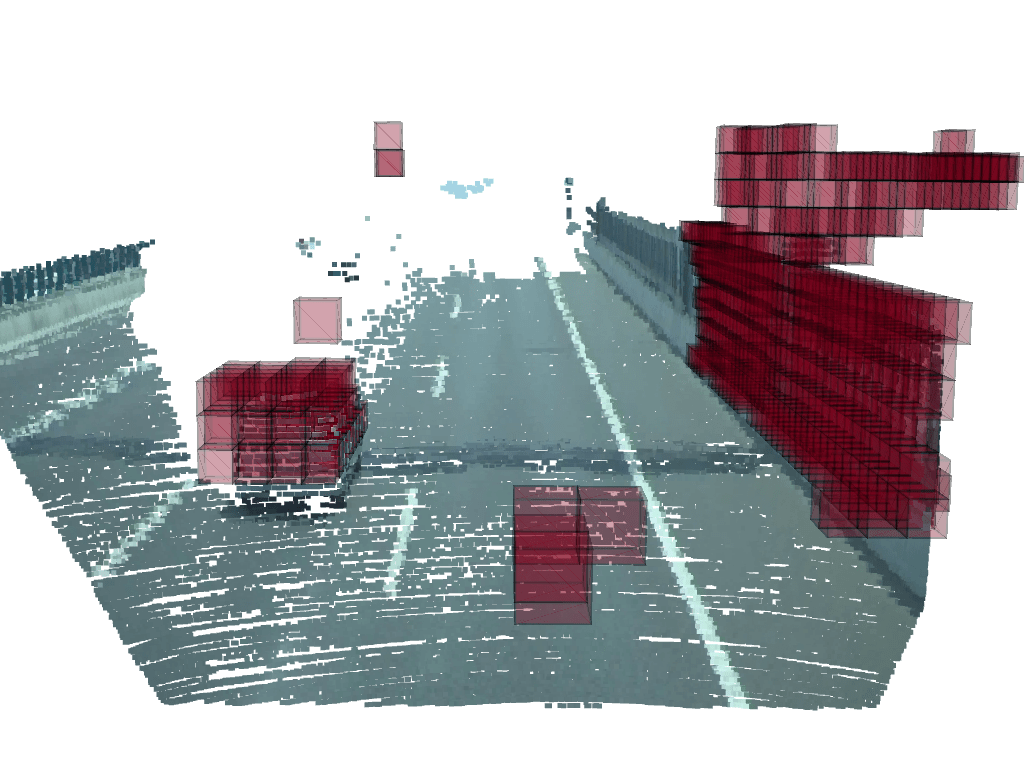}
    \end{subfigure}
    \begin{subfigure}{.19\linewidth}
        \centering
        \includegraphics[width=\linewidth, height=2.2cm]{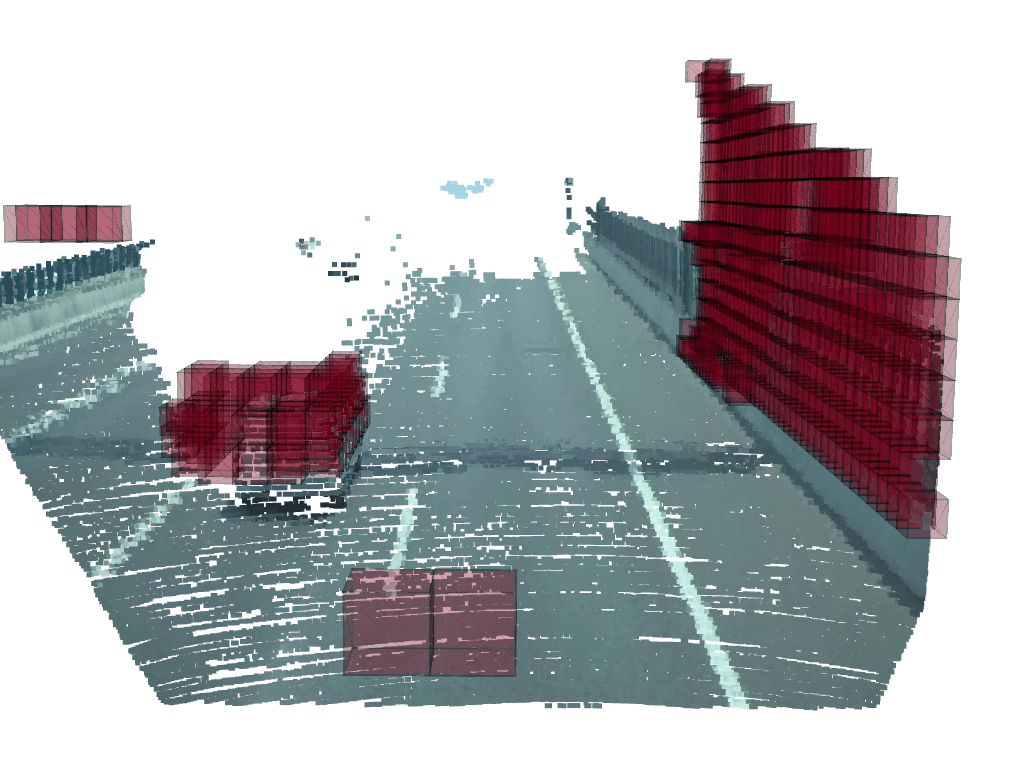}
    \end{subfigure}
    \begin{subfigure}{.19\linewidth}
        \centering
        \includegraphics[width=\linewidth, height=2.2cm]{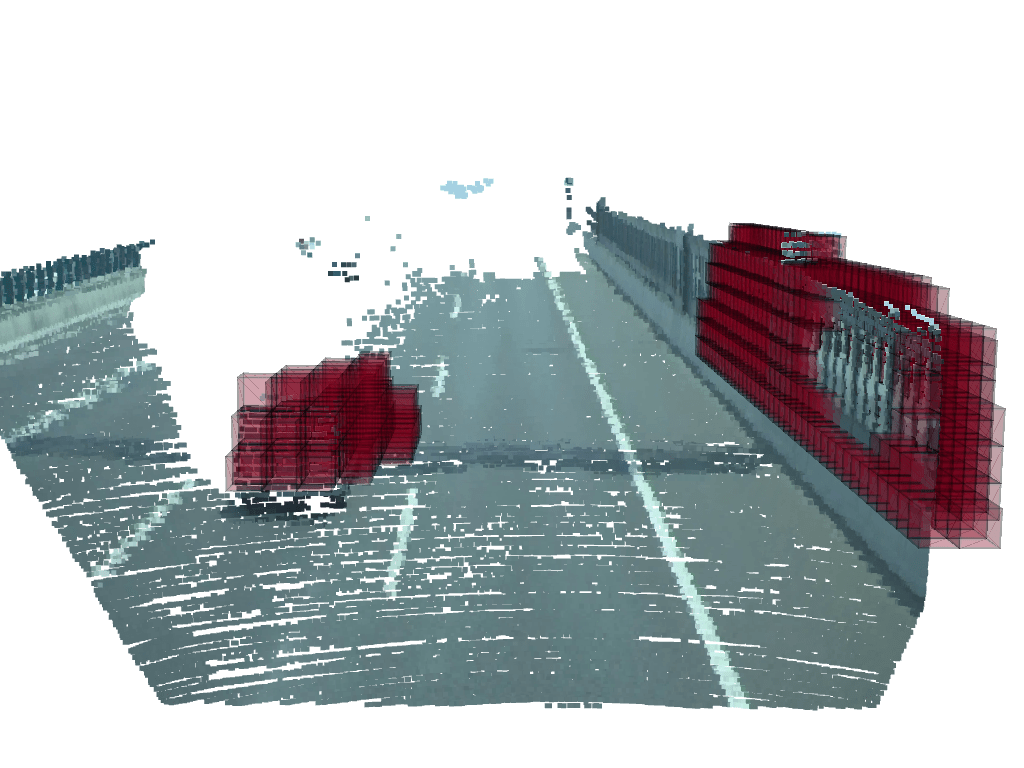}
    \end{subfigure}
    
    \begin{subfigure}{.19\linewidth}
        \centering
        \includegraphics[width=\linewidth, height=2.2cm]{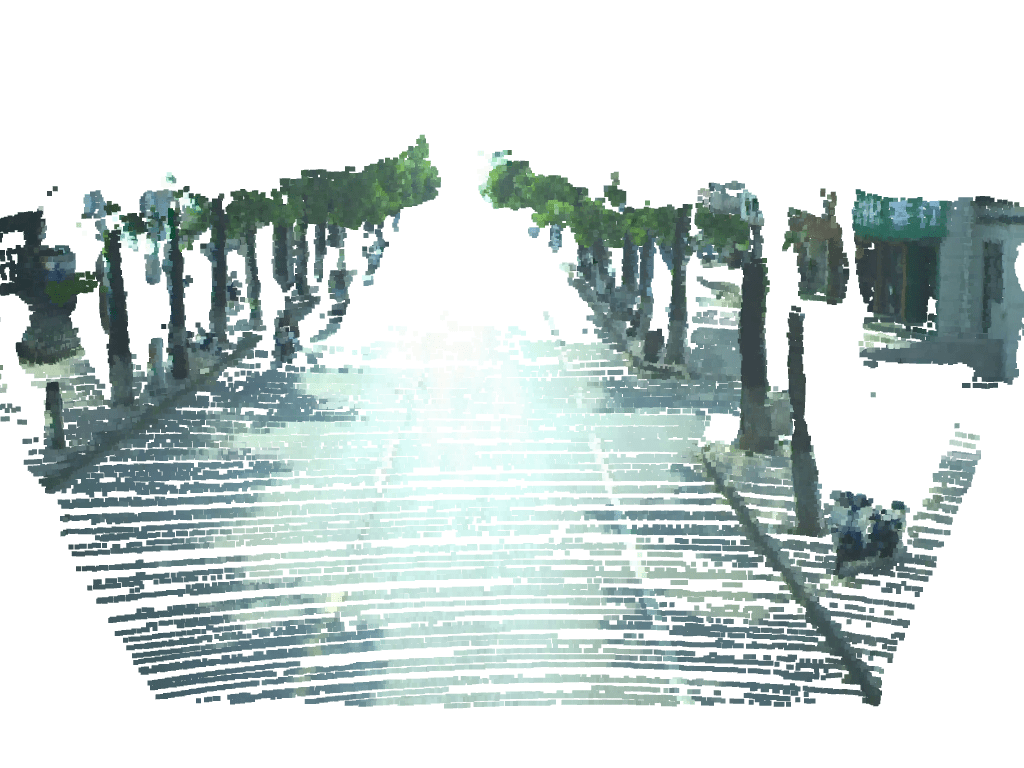}
    \end{subfigure}
    \begin{subfigure}{.19\linewidth}
        \centering
        \includegraphics[width=\linewidth, height=2.2cm]{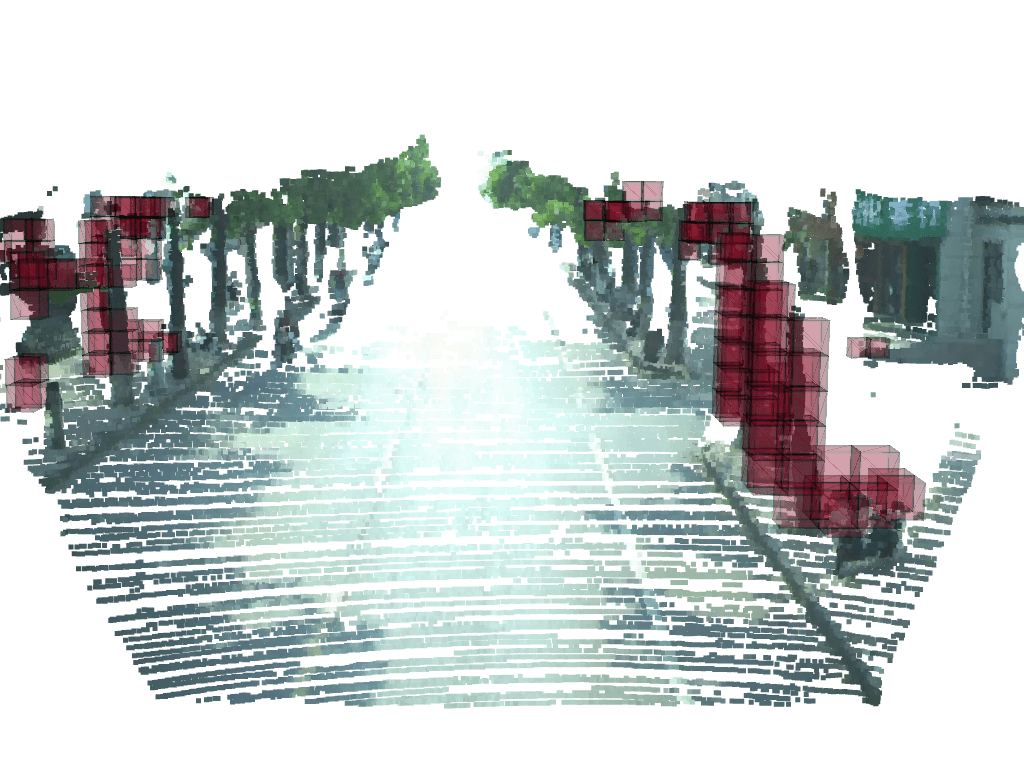}
    \end{subfigure}
    \begin{subfigure}{.19\linewidth}
        \centering
        \includegraphics[width=\linewidth, height=2.2cm]{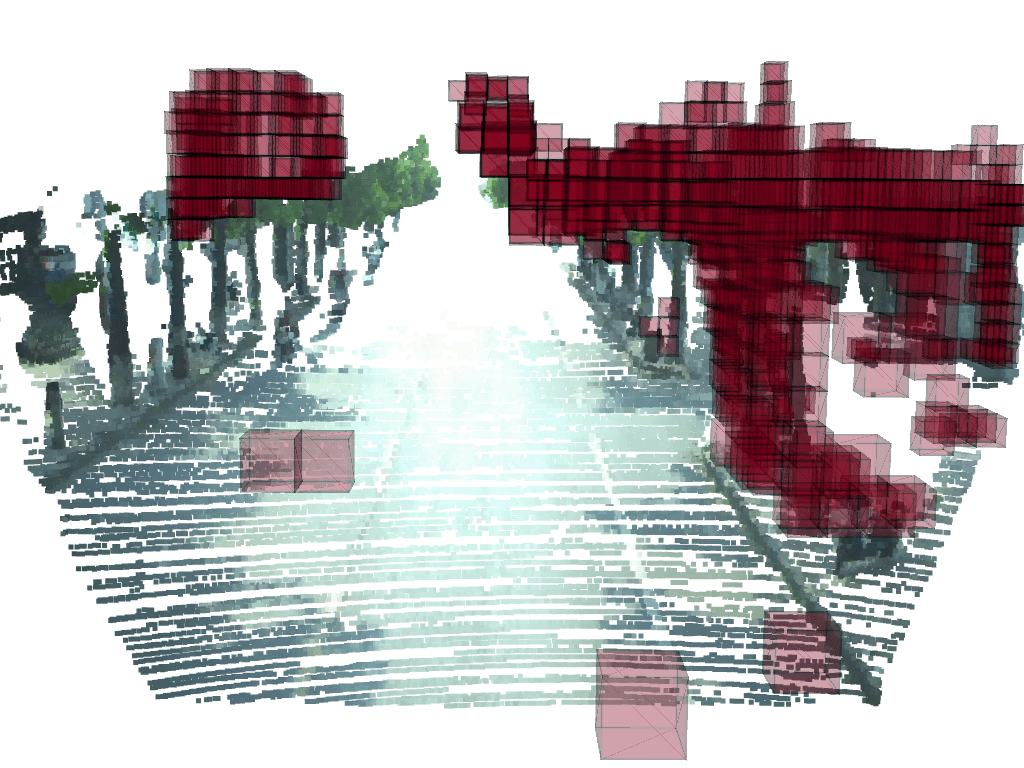}
    \end{subfigure}
    \begin{subfigure}{.19\linewidth}
        \centering
        \includegraphics[width=\linewidth, height=2.2cm]{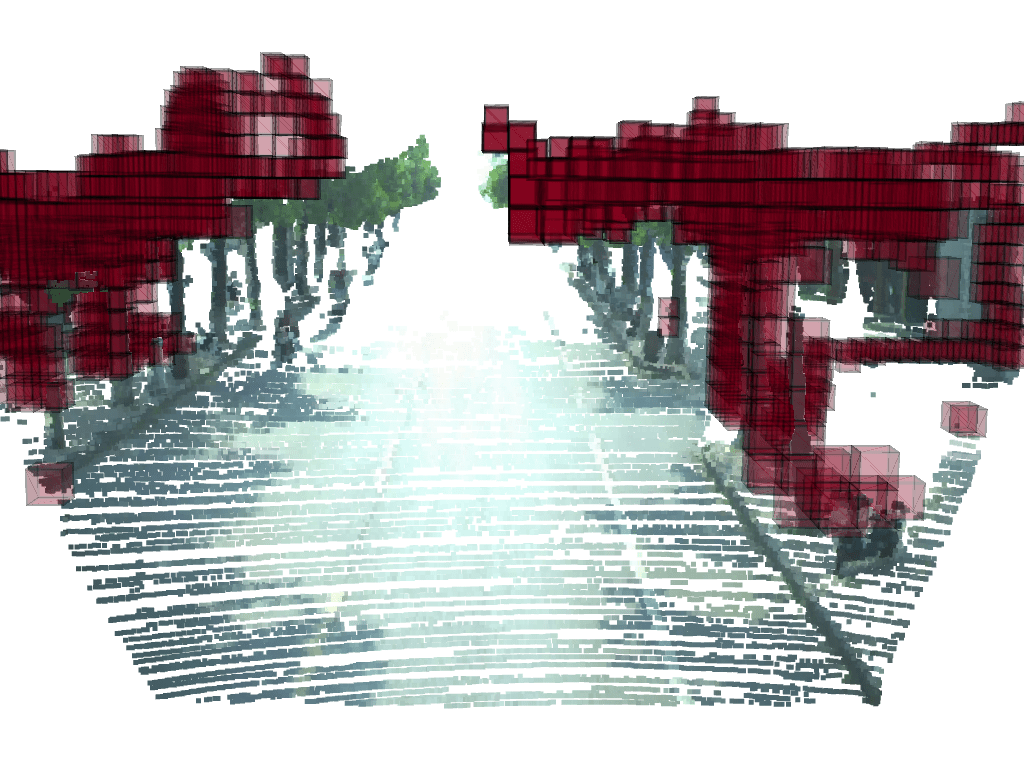}
    \end{subfigure}
    \begin{subfigure}{.19\linewidth}
        \centering
        \includegraphics[width=\linewidth, height=2.2cm]{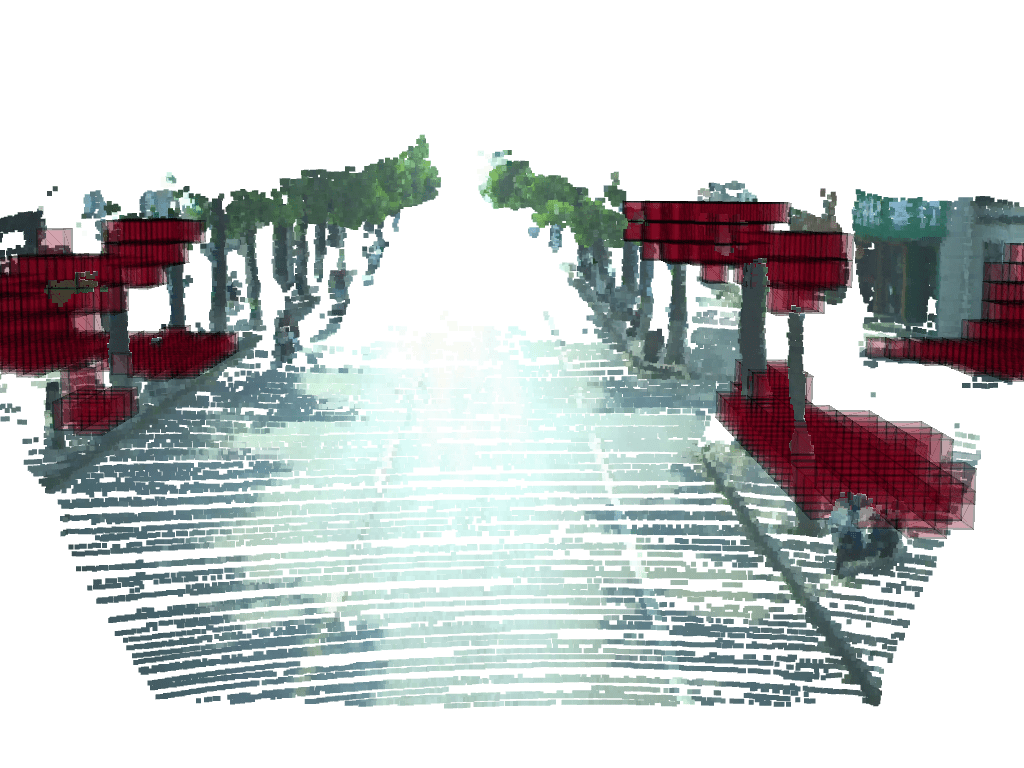}
    \end{subfigure}

      \begin{subfigure}{.19\linewidth}
        \centering
        \includegraphics[width=\linewidth, height=2.2cm]{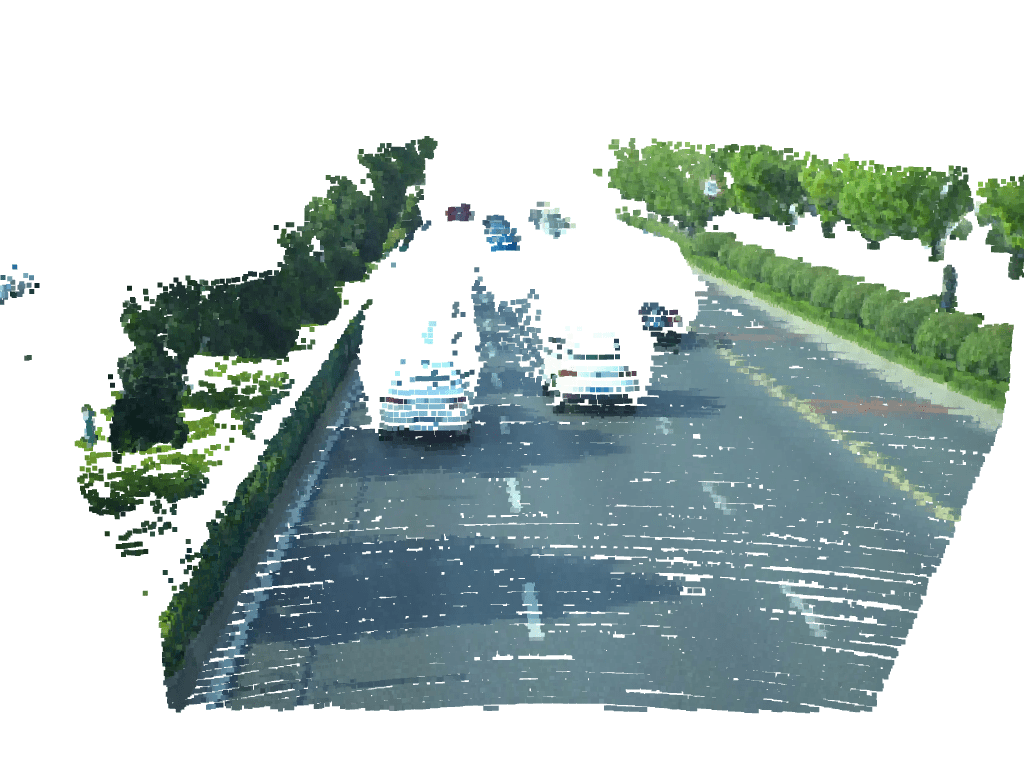}
        \caption*{Background}
    \end{subfigure}
    \begin{subfigure}{.19\linewidth}
        \centering
        \includegraphics[width=\linewidth, height=2.2cm]{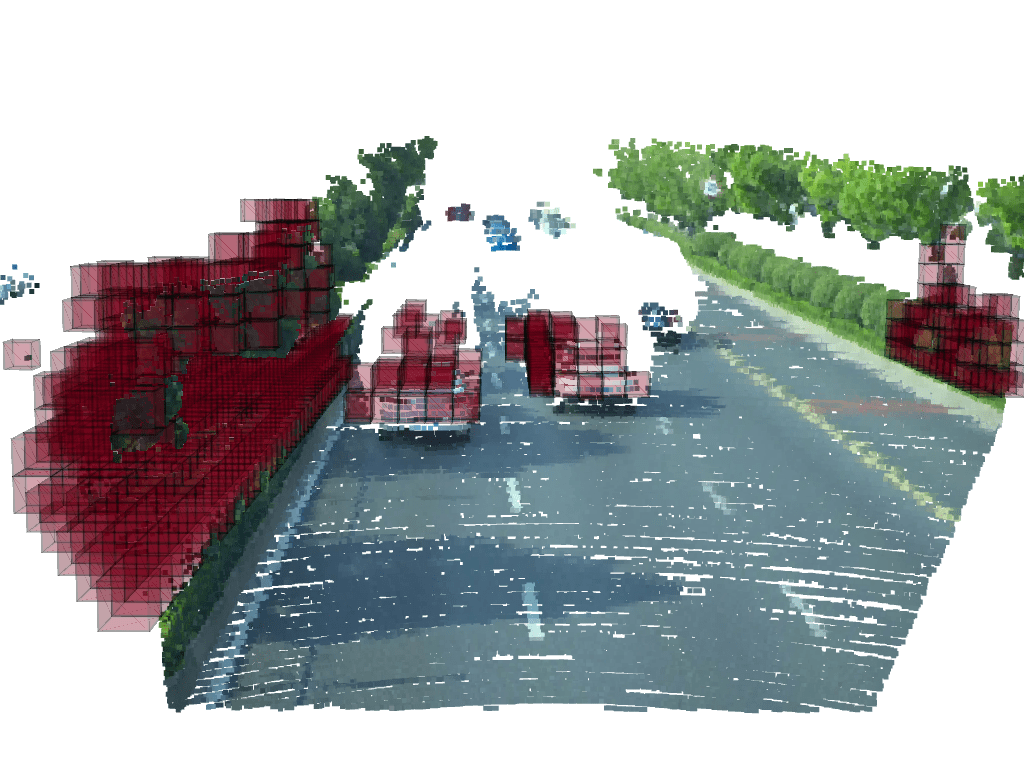}
        \caption*{Ground Truth}
    \end{subfigure}
    \begin{subfigure}{.19\linewidth}
        \centering
        \includegraphics[width=\linewidth, height=2.2cm]{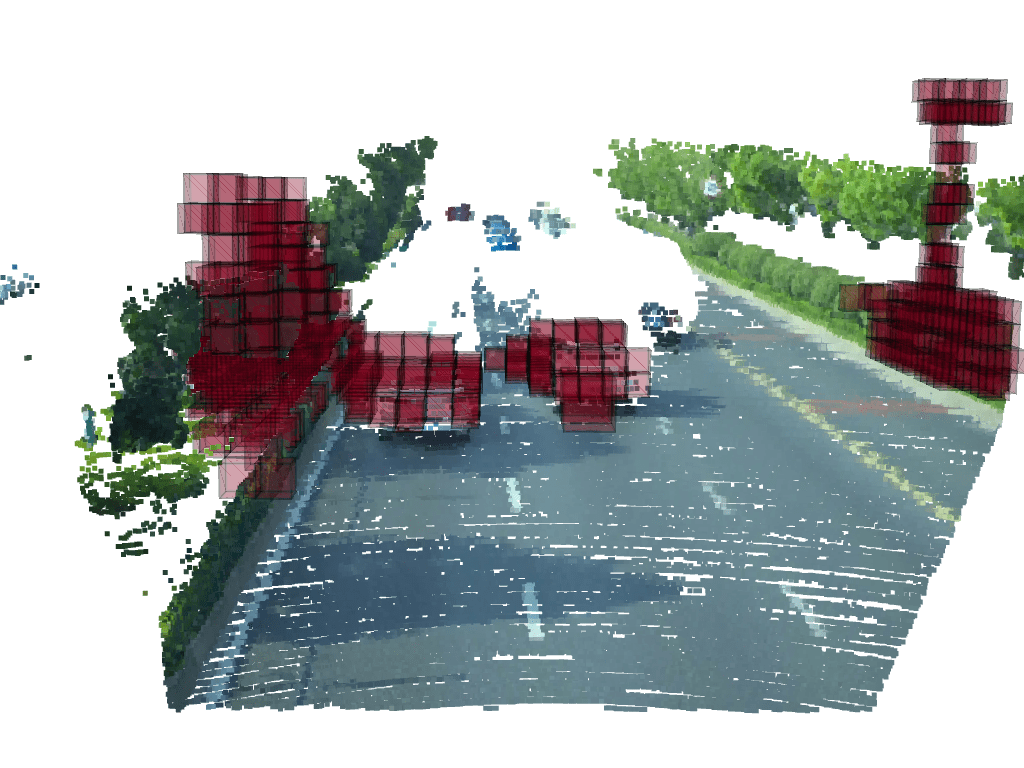}
        \caption*{SGBM \cite{sgm}}
    \end{subfigure}
    \begin{subfigure}{.19\linewidth}
        \centering
        \includegraphics[width=\linewidth, height=2.2cm]{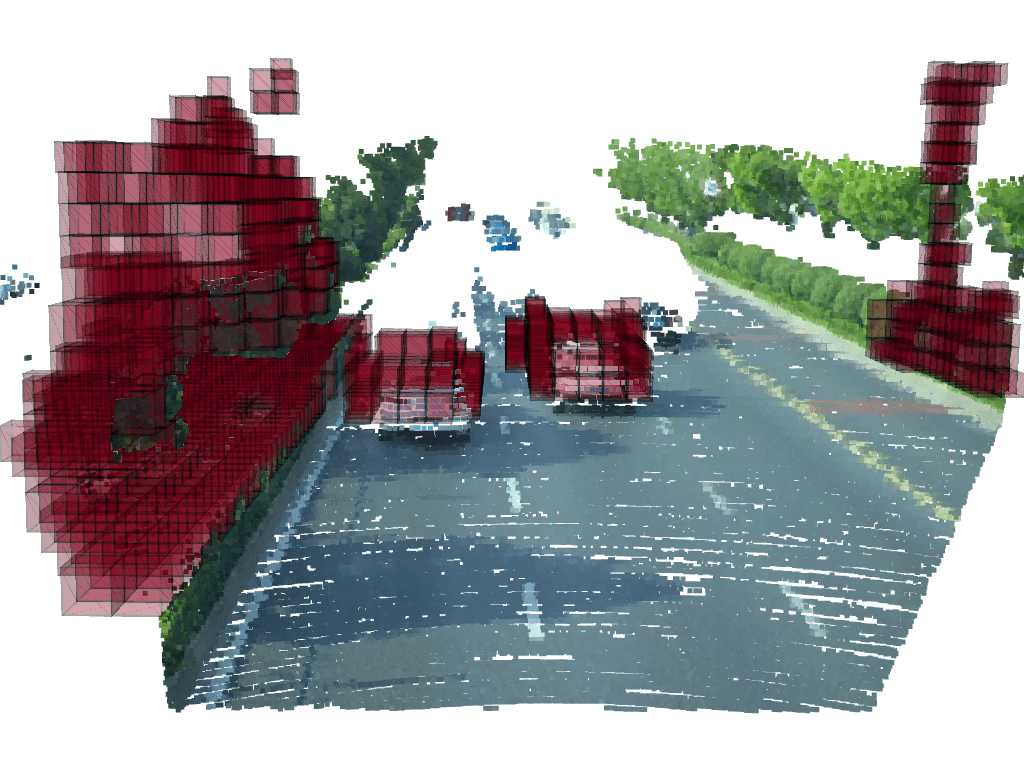}
        \caption*{Lac-GwcNet \cite{liu_local_2022}}
    \end{subfigure}
    \begin{subfigure}{.19\linewidth}
        \centering
        \includegraphics[width=\linewidth, height=2.2cm]{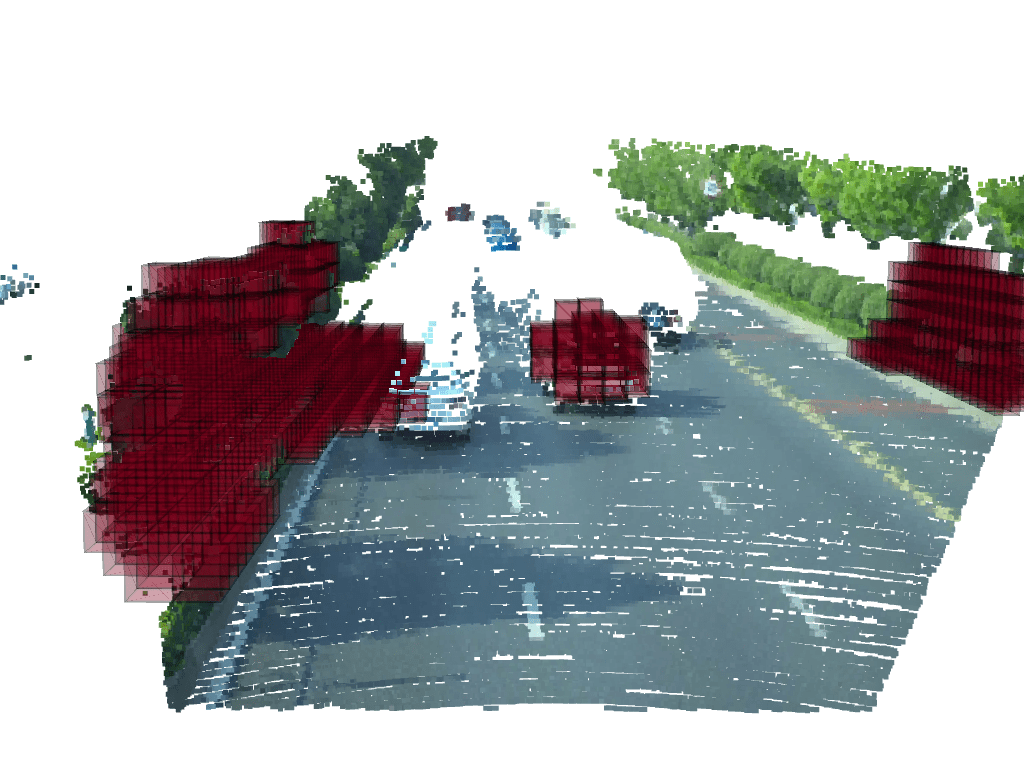}
        \caption*{\textbf{Ours}}
    \end{subfigure}

    
    \caption{Results for DrivingStereo dataset. Output with $64^3$ resolution is shown above.
    }
    \label{fig:qualitative-result}
\end{figure*}

\section{Experiments}
\label{sec:experiment}

In this section, we evaluate the performance and efficiency of the proposed obstacle detection solution. We structure our evaluation into two parts: public dataset evaluation and real-world experiments. We also include ablation studies where we examine the effectiveness of our proposed \CostVolumeName{} and octree decoder structure.

We use DrivingStereo \cite{yang2019drivingstereo} to train and evaluate our model. Similar to the widely used KITTI datasets\cite{KITTI2012, KITTI2015}, DrivingStereo is a driving dataset but with more samples. 

We set $l_v=0.5$, $n_x=n_y=n_z=64$. In other words, our ROI is a $64^3$ cubic volume in front of us with 32 meters distance, 16 meters left, 16 meters right, and 32 meters towards the top. In practice, we set the size of the latent vector to $n_{latent}=128$.
Our model is trained for 30 epochs using an Adam optimizer \cite{kingma_adam_2017} with a learning rate of 0.001, $\beta_1=0.9, \beta_2=0.999$. We form a batch size of 16 stereo pairs. To implement the sparse convolution in our octree decoder, we use the \emph{SpConv}\footnote{https://github.com/traveller59/spconv} library, the submanifold sparse convolution layer \cite{graham2017submanifold, graham20183d}, and PyTorch\cite{NEURIPS2019_9015}.

\subsection{Quantitative Results}

Similar to 3D reconstruction tasks\cite{10.1007/978-3-030-11009-3_37} \cite{8890921}, our final output is a voxel occupancy grid. Therefore, we choose IoU and Chamfer Distance (CD) as our evaluation metrics. Besides the accuracy metrics, we also consider computation cost metrics: multiply-accumulate (MACs) and model parameters. Models with more significant MACs typically demand more computation power and process slower.



\subsubsection{\titlecap{\CostVolumeName{}}}

We perform an ablation study to evaluate the effectiveness of our proposed \CostVolumeName{} and its impact on the entire network. We compare it with the other three types of cost volume (Table \ref{tab:cost_volume}). The original interlacing cost volume (ICV)\cite{shamsafar2022mobilestereonet} has 48 disparity levels, and the simplified version of it where we only extract the even number of disparities, results in 24 levels of disparity (Even). We also obtained a further simplified version with only 12 levels of disparity (same as our \CostVolumeName{}) using the same strategy. 

\begin{table}[h]
\vspace*{0.15cm}

    \centering
    \resizebox{\columnwidth}{!}{%
    \begin{tabular}{l r r r r}
    \hline
    Cost Volume Type & CD $\downarrow$ & IoU $\uparrow$ & MACs \tablefootnote{MACs based on $3\times400\times880$ RGB stereo pair} $\downarrow$ & Parameters $\downarrow$\\
    \hline
    48 Levels (ICV)  & 3.56 & 0.32 & 29.01G & 5.49M \\
    24 Levels (Even) & \underline{2.96} & \underline{0.34}  & 21.00G & 5.43M \\
    12 Levels (Even) & 3.05 & 0.33  & \textbf{16.99G} & \textbf{5.40M} \\
    \hline
    12 Levels (\textbf{Ours}) & \textbf{2.40} & \textbf{0.35} & \underline{17.31G} & \textbf{5.40M} \\
    \hline
    \end{tabular}
    }
    \caption{Numerical results on the DrivingStereo testing set. The best result is bold, and the second-best result is underlined.}
    \label{tab:cost_volume}
\end{table}

Our \CostVolumeName{} achieves leading performance with computation cost close to the minimum. Compared to the interlacing cost volume that has the 24 disparity levels, our approach achieves 19\% lower CD and 18\% lower computation cost. Therefore, we show the effectiveness of cost volume optimization based on ROI.

\subsubsection{Octree Pruning}
In this section, we demonstrate the efficacy of our octree structure pruning strategy by comparing it with the same network using dense grid implementation (Dense). Following the training schemes in \cite{Chitta_2020_WACV}, we choose the active sites at each octree level using both ground truth (GT) and predicted label (Pred). In addition, we evaluate the baseline network performance that only produces a single output at the end of the network (Straight). Our results are summarized in Table \ref{tab:octree-result}.
\begin{table}[!ht]
    \centering
    \resizebox{\columnwidth}{!}{%
    \begin{tabular}{lrrrrrr}
    \hline
    Method & Level & CD $\downarrow$ & IoU $\uparrow$ & MACs $\downarrow$ & Parameters $\downarrow$\\ 
    \hline
    Straight & 4 & 4.89 & \textbf{0.38} & 17.31G & 5.40M \\
    \hline
    \multirow{4}{*}{Dense}  & 1 & \textbf{5.68} & \textbf{0.78} & 15.05G & \multirow{4}{*}{5.40M}      \\
                            & 2 & \textbf{3.62} & \textbf{0.64} &  15.72G  &                            \\
                            & 3 & \textbf{2.68} & \textbf{0.50} &  16.86G  &                            \\
                            & 4 & \textbf{2.40} & \underline{0.35} & 17.31G   &                            \\
    \hline
    \multirow{4}{*}{Sparse-GT}  & 1 & 6.75 & 0.75 & \multirow{4}{*}{\textbf{14.94G}} & \multirow{4}{*}{5.40M}      \\
                            & 2 & 4.16 & 0.58 &    &                            \\
                            & 3 & 3.05 & 0.42 &    &                            \\
                            & 4 & 2.56 & 0.27 &    &                            \\
    \hline
    \multirow{4}{*}{Sparse-Pred}  & 1 & \underline{5.84} & \textbf{0.78} & \multirow{4}{*}{\textbf{14.94G}} & \multirow{4}{*}{5.40M}      \\
                            & 2 & \underline{3.69} & \textbf{0.64} &    &                            \\
                            & 3 & \underline{2.75} & \textbf{0.50} &    &                            \\
                            & 4 & \underline{2.54} & \underline{0.35} &    &                            \\
    \hline
    \end{tabular}
    }
    \caption{Our octree-pruned model (Sparse-Pred) achieves nearly the same performance as the dense grid network while reducing the computation cost.}
    \label{tab:octree-result}
\end{table}

From the experimental result, we conclude that the octree decoder achieves nearly the same performance as the dense grid decoder, while further reducing the computation cost by 13.2\%.

\subsubsection{\NetworkName{}}
We compare the performance of the \NetworkName{} using \CostVolumeName{} with several state-of-the-art stereo deep learning models and the traditional SGBM algorithm in Table \ref{tab:all-ds}. 

We apply the standard pipeline for obstacle detection using a stereo camera \cite{9340699}, where the disparity estimation is transformed into the depth map, the point cloud, and ultimately the voxel grid.

\begin{table}[!ht]
    \centering
    \resizebox{\columnwidth}{!}{%
    \begin{tabular}{lrrrrr}
    \hline
    Method & CD $\downarrow$ & IoU $\uparrow$ & MACs $\downarrow$ & Parameters $\downarrow$ & Runtime $\downarrow$\\
    \hline
    MSNet (2D) \cite{shamsafar2022mobilestereonet} & 17.03 & 0.15 & 91.04G & 2.35M & 2.61s \\
    
    MSNet (3D) \cite{shamsafar2022mobilestereonet} & 12.2 & 0.21 & 409.09G & \underline{1.77M} &  \\
   
    ACVNet \cite{xu2022ACVNet} & 11.36 & 0.22 & 644.57G & 6.17M & 4.38s \\
    
    Lac-GwcNet \cite{liu_local_2022} & \underline{6.22} & \underline{0.33} & 775.56G & 9.37M & 5.83s \\
    
    CFNet \cite{Shen_2021_CVPR} \tablefootnote{Due to CFNet input requirement, MACs based on $3\times384\times864$}  & 7.86 & 0.22 & 428.40G & 22.24M &  \\
   
    SGBM \cite{sgm} \tablefootnote{Implemented by OpenCV(StereoSGBM)} & 16.59 & 0.25 & \textbf{N/A} & \textbf{N/A} & \textbf{0.08s} \\
    \hline
    \textbf{Ours} & \textbf{2.40} & \textbf{0.35} & \underline{17.31G} & 5.40M & \underline{0.55s} \\
    \hline
    \end{tabular}
    }
    \caption{Numerical results on the DrivingStereo dataset. We use Intel i7-12700K runtime to compare all approaches.}
    \label{tab:all-ds}
\end{table}

Our method outperforms the compared approaches by a large margin in terms of accuracy and speed. We show the possibility of applying a deep learning stereo model in the obstacle avoidance task. However, comparing with SGBM in terms of \emph{CPU runtime} remains a challenge. Our model runs at \textbf{35Hz} on RTX 3090 GPU without any optimization.


\subsection{Qualitative Results}

We present the qualitative result in Fig. \ref{fig:qualitative-result} in which we compare the result with SGBM and the state-of-the-art model Lac-GwcNet\cite{liu_local_2022}. The red voxels represent the detected obstacles. Our method produces the least noise and the closest result to the ground truth. We achieve similar or better results than Lac-GwcNet with only 2\% computation cost.

\subsection{Real-World Experiments}

\begin{figure}[ht!]
    \centering
    \begin{subfigure}{.45\linewidth}
        \centering
        \includegraphics[width=\linewidth, height=3.0cm]{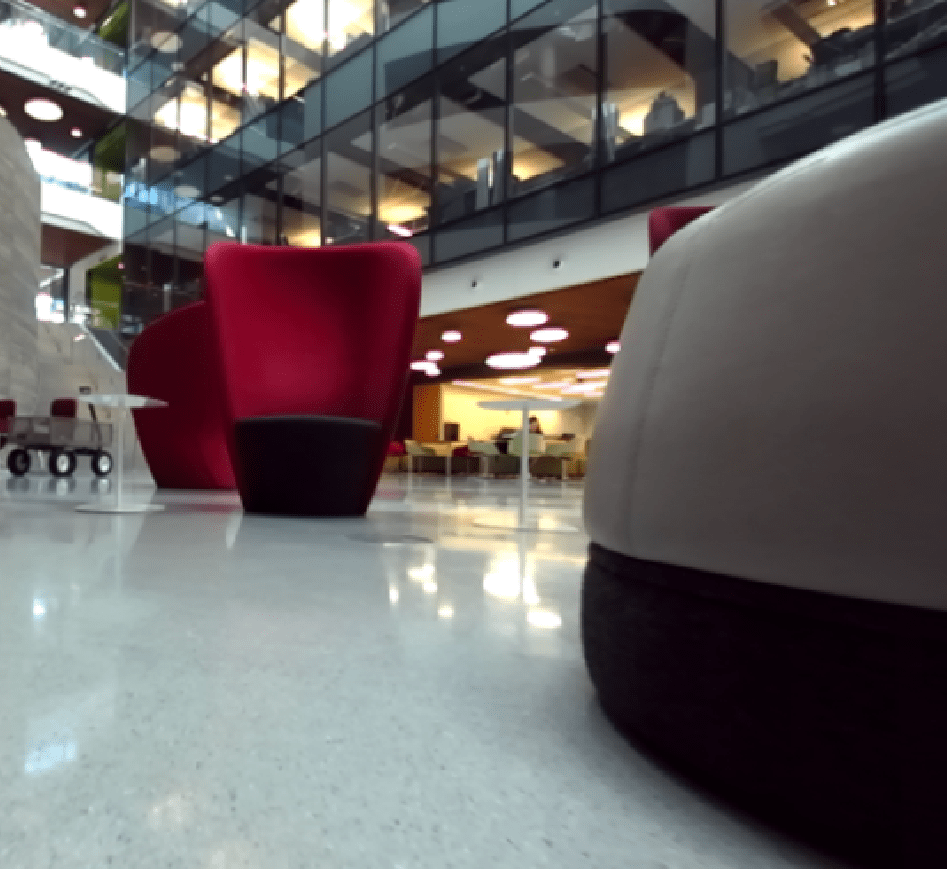}
    \end{subfigure}
    \hspace{-0.1cm}
    \begin{subfigure}{.45\linewidth}
        \centering
        \includegraphics[width=\linewidth, height=3.0cm]{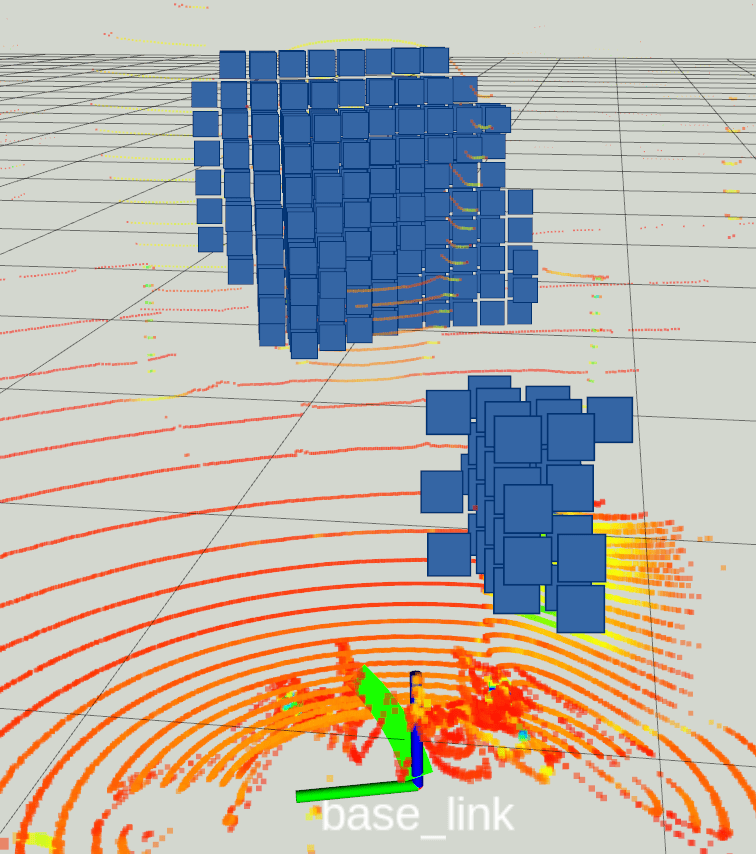}
    \end{subfigure}
    \caption{We experiment in an unstructured environment with no prior knowledge. The robot is given a goal across the environment and avoids obstacles during navigation.}
    \label{fig:real_world_exp}
\end{figure}

We adopt an autonomous navigation method with only a local planner from the \emph{move\_base} of ROS Noetic on Ubuntu 20.04. The LiDAR is only used for visualization purposes during the experiment. Our proposed method is the only input to the navigation stack that operates on Jackal's onboard computer (NVIDIA Jetson TX2). The goals are specified in the environment (Fig. \ref{fig:real_world_exp}). Although the robot reaches its destination successfully, we observe that noise is often generated by the ground reflection, leading to failed attempts. Due to the absence of a global planner, the robot is likely to get stuck at local minima. We refer readers to our supplementary video for a demonstration of the experiments.


\section{Conclusion}
\label{sec:conclusion}

This paper presented a novel obstacle detection network based on the stereo vision that can run in real-time on an onboard computer. We proposed voxel cost volume to keep the computation within our ROI and match the disparity estimation resolution with the occupancy grid. We leveraged the octree-generating decoder to prune the insignificant computation on irrelevant spaces. Furthermore, we integrated the hierarchical output with the navigation framework to provide adaptive resolution while navigating. To support the data-based robot applications, we collected and released a high-quality real-world indoor/outdoor stereo dataset via robot teleoperation. Our work shows the possibility of utilizing a deep stereo model in the obstacle detection task.






\section*{ACKNOWLEDGMENT}

This work was completed in part using the Discovery cluster, supported by Northeastern University’s Research Computing team. The authors would like to thank Ying Wang, Nathaniel Hanson, Mingxi Jia, and Chenghao Wang for their help and discussions.

\bibliographystyle{IEEEtran}
\bibliography{root, zotero}

\end{document}